\documentclass[11pt]{article}
\usepackage{acl-hlt2011}

\usepackage{amsmath} 
\usepackage{stmaryrd}
\usepackage{latexsym} 
\usepackage{epsf} 
\usepackage[all]{xy}  
\usepackage{lscape} 
\usepackage{verbatim}    
\usepackage{times}  
\usepackage{color}
\usepackage{epsfig}
\usepackage{qtree} 
\usepackage{multirow}
\usepackage{url}

\title{Experimental Support for a Categorical Compositional\\Distributional Model of Meaning}

\author{Edward Grefenstette \\
  University of Oxford\\ 
  Department of Computer Science \\
  Wolfson Building, Parks Road \\
  Oxford OX1 3QD, UK \\
  {\tt edward.grefenstette@cs.ox.ac.uk} \\\And
  Mehrnoosh Sadrzadeh \\
  University of Oxford\\ 
  Department of Computer Science \\
  Wolfson Building, Parks Road \\
  Oxford OX1 3QD, UK \\
  {\tt mehrs@cs.ox.ac.uk} \\}

\date{}
\begin{document}

\maketitle


\begin{abstract}
Modelling compositional  meaning for sentences using empirical distributional methods has been a challenge for computational linguists. We implement the abstract categorical model of~\newcite{Coeckeetal} using data from the BNC and evaluate it. The implementation is based on unsupervised learning of matrices for relational words and applying them to the vectors of their arguments. The evaluation is based on the word disambiguation task developed by~\newcite{Lapata} for intransitive sentences, and on a similar new experiment designed for transitive sentences. Our model matches the results of its competitors in the first experiment, and betters them in the second. The general improvement in results with increase in syntactic complexity showcases the compositional power of our model.
\end{abstract}

\section{Introduction}
As competent language speakers, we humans can almost trivially make sense of sentences we've never seen or heard before. We are naturally good at understanding ambiguous words given a context, and forming the meaning of a sentence from the meaning of its parts. But while human beings seem comfortable doing this, machines fail to deliver. Search engines such as Google either fall back on bag of words models---ignoring syntax and lexical relations---or exploit superficial models of lexical semantics to retrieve pages with terms related to those in the query~\cite{Manning}. 

However, such models fail to shine when it comes to processing the semantics of phrases and sentences. Discovering the process of meaning assignment in natural language is among the most challenging and foundational questions of linguistics and computer science. The findings thereof will increase our understanding of cognition and intelligence and shall assist in applications to automating language-related tasks such as  document search.  

Compositional type-logical approaches~\cite{Montague,Lambek} and distributional models of lexical semantics~\cite{Schutze,Firth} have provided two partial orthogonal solutions to the question.  Compositional formal semantic models stem from classical ideas from mathematical logic, mainly Frege's principle that the meaning of a sentence is a function of the meaning of its parts~\cite{Frege}. Distributional models are more recent and can be related to Wittgenstein's later philosophy of `meaning is use', whereby  meanings of  words can be determined from their context~\cite{Wittgenstein}.  The logical models relate to well known and robust logical formalisms, hence offering a scalable theory of meaning which can be used to reason inferentially. The distributional models have found their way into real world applications such as thesaurus extraction~\cite{Grefenstette,Curran} or automated essay marking~\cite{Landauer}, and have connections to semantically motivated information retrieval~\cite{Manning}. This two-sortedness of defining properties of meaning:  `logical form' versus `contextual use',  has left the quest for  `what is the foundational structure of meaning?' even more of a challenge.

Recently,~\newcite{Coeckeetal} used high level cross-disciplinary techniques from logic, category theory, and physics  to bring the above two approaches together. They developed a unified mathematical framework whereby a sentence vector is by definition  a function of the Kronecker product of its word vectors. A concrete instantiation of this theory was exemplified on  a toy hand crafted corpus by~\newcite{Grefenetal}.  In this paper we  implement it by training the model over the entire BNC. The highlight of our implementation is that words with relational types,  such as verbs, adjectives, and adverbs  are matrices that \emph{act} on their arguments. We provide  a general algorithm for building (or indeed learning) these matrices from the corpus.  

The implementation is evaluated against the task provided by \newcite{Lapata} for disambiguating intransitive verbs, as well as a similar new experiment for  transitive verbs.  Our model improves on the best method evaluated in \newcite{Lapata} and offers promising results for the transitive case, demonstrating its scalability in comparison to that of other models.  But we still feel there is need for a different class of experiments to showcase merits of compositionality in a statistically significant manner.  Our work shows   that the categorical compositional distributional model of meaning permits a practical implementation and that this opens the way to the production of large scale compositional models. 


\section{Two Orthogonal Semantic Models}

\paragraph{Formal Semantics}
To compute the meaning of a sentence consisting of $n$ words, meanings of these  words must \emph{interact} with one another.  
In formal semantics, this further interaction  is  represented as a function derived from the grammatical structure of the sentence, but meanings of words are  amorphous objects of the domain: no distinction is  made between words that have the same type. Such models consist of a pairing of syntactic interpretation rules (in the form of a grammar) with semantic interpretation rules, as exemplified by the simple model presented in Figure \ref{fig:formal_semantic_model}.

\begin{figure}[h]
\begin{center}
		\small
		\begin{tabular}{l|l}
			Syntactic Analysis & Semantic Interpretation\\
			\hline
			S $\to$ NP VP & $|VP|(|NP|)$\\
			NP $\to$ cats, milk, etc. & $|\textrm{cats}|,\,|\textrm{milk}|,\,\ldots$\\
			VP $\to$ Vt NP & $|Vt|(|NP|)$\\
			Vt $\to$ like, hug, etc. & $\lambda yx.|\textrm{like}|(x,y),\,\ldots$ \\
		\end{tabular}
\end{center}
\caption{A simple model of formal semantics.}
\label{fig:formal_semantic_model}
\end{figure}

\begin{figure}[h]
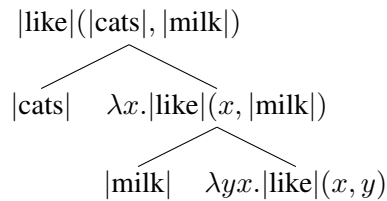

\Tree [.$|\textrm{like}|(|\textrm{cats}|,|\textrm{milk}|)$ $|\textrm{cats}|$ [.$\lambda x.|\textrm{like}|(x,|\textrm{milk}|)$   $|\textrm{milk}|$ $\lambda yx.|\textrm{like}|(x,y)$ ] ]
\caption{A parse tree showing a semantic derivation.}
\label{fig:semantic_parse}
\end{figure}

The parse of a sentence such as ``cats like milk'' typically produces its semantic interpretation by substituting semantic representation  for their grammatical constituents and applying $\beta$-reduction where needed. Such a derivation is shown in Figure \ref{fig:semantic_parse}. 

This methodology is used to  translate sentences of natural language into logical formulae, then use computer-aided automation tools to reason about them~\cite{Alshawi}. One  major  drawback is that the result of such analysis can only deal with truth or falsity as the meaning of a sentence, and says nothing about the closeness in meaning or topic of expressions beyond their truth-conditions and what models satisfy them, hence do not perform well on language tasks such as search. Furthermore, an underlying domain of objects and a valuation function must be provided, as with any logic, leaving open the question of how we might \emph{learn} the meaning of language using such a model, rather than just use it.

\paragraph{Distributional Models}

Distributional models of semantics, on the other hand,  dismiss the interaction between syntactically linked words and are solely concerned with lexical semantics. Word meaning is obtained empirically by examining the \emph{contexts}\footnote{\emph{E.g.}~words which appear in the same sentence or $n$-word window, or words which hold particular grammatical or dependency relations to the word being learned.} in which a word appears, and equating the meaning of a word with the distribution of contexts it shares. The intuition is that context of use is what we appeal to in learning the meaning of a word, and that words that frequently have the same sort of context in common are likely to be semantically related.

For instance, beer and sherry are both drinks, alcoholic, and often cause a hangover. We expect these facts to be reflected in a sufficiently large corpus: the words `beer' and `sherry' occur within the context of identifying words such as `drink',  `alcoholic' and `hangover' more frequently than they occur with other content words. 

Such context distributions can be encoded as vectors in a high dimensional space with contexts as basis vectors. For any word vector $\overrightarrow{\text{word}}$, the scalar weight $c^{word}_i$ associated with each context basis vector $\overrightarrow{n_i}$ is a function of the number of times the word has appeared in that context. Semantic vectors $(c^{word}_1, c^{word}_2, \cdots, c^{word}_n)$ are also denoted  by sums of such weight/basis vector pairs:

\[
\overrightarrow{\text{word}} = \sum_i{c^{word}_i \overrightarrow{n_i}}
\]

\noindent
Learning a semantic vector is just learning its basis weights from the corpus. This setting offers geometric means to reason about semantic similarity (\emph{e.g.}~via cosine measure or $k$-means clustering), as discussed in~\newcite{Widdows}.

The principal drawback of such models is their non-compositional nature: they ignore grammatical structure and logical words, and hence cannot  compute the meanings of phrases and sentences in the same efficient way that they do for words. Common operations discussed in \cite{Lapata} such as vector addition ($+$) and component-wise multiplication ($\odot$, \emph{cf.}~$\S$\ref{sec:building} for details) are commutative, hence if $\overrightarrow{vw} = \overrightarrow{v} + \overrightarrow{w}$ or $\overrightarrow{v} \odot \overrightarrow{w}$, then $\overrightarrow{vw} = \overrightarrow{wv}$, leading to unwelcome equalities such as 
\[
\overrightarrow{\textrm{the dog bit the man}} = \overrightarrow{\textrm{the man bit the dog}}
\]
Non-commutative operations, such as the Kronecker product (\emph{cf.}~$\S$\ref{sec:building} for definition) can take word-order into account~\cite{Smolensky} or even some more complex syntactic relations, as described in \newcite{ClarkPulman}. However, the dimensionality of sentence vectors produced in this manner differs for sentences of different length, barring all sentences from being compared in the same vector space, and growing exponentially with sentence length hence quickly becoming computationally intractable.

\section{A Hybrid Logico-Distributional Model}
Whereas semantic compositional mechanisms for set-theoretic constructions are well understood, there are no obvious corresponding methods for vector spaces.  To solve this problem, ~\newcite{Coeckeetal} use the abstract setting of  category theory  to turn the grammatical structure of a sentence into a morphism compatible with the higher level logical structure of vector spaces.  

One pragmatic consequence of this abstract idea is as follows.  In distributional models,  there is a meaning vector for each word, \emph{e.g.}~$\overrightarrow{\text{cats}}$, $\overrightarrow{\text{like}}$, and $\overrightarrow{\text{milk}}$. The  logical recipe tells us to \emph{apply} the meaning of the verb to the meanings of subject and object. But how can a vector \emph{apply} to other vectors? The solution proposed above implies that one needs to  have different levels of meaning for words with different types. This is similar to logical models where verbs are relations and  nouns are atomic sets. So verb vectors should be built differently from noun vectors, for instance as matrices. 

The general information as to which words should be matrices and which words atomic vectors is in fact encoded in the type-logical representation of the grammatical structure of the sentence. This is the linear map with word vectors as input and sentence vectors as output.  Hence, at least theoretically, one should be able to build sentence vectors and compare their synonymity in exactly the same way as one measures word synonymity. 

\paragraph{Pregroup Grammars}

The aforementioned linear maps turn out to be the grammatical reductions of a type-logic called a Lambek pregroup grammar~\cite{Lambek}\footnote{The usage of pregroup types is not essential, the types of any other logic, for instance CCG can be used, but should be translated into the language of pregroups.  
}.  Pregroups and vector spaces share the same high level mathematical structure, referred to as a \emph{compact closed category}, for a proof and details of this claim see~\newcite{Coeckeetal};  for a friendly introduction to category theory, see~\newcite{Coecke}. One consequence of this parity is that the   grammatical reductions  of a pregroup grammar can be directly transformed into  linear maps that act on vectors.

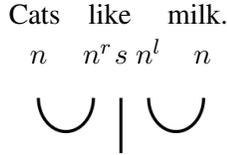
\begin{figure}[h]
\begin{center}
\begin{minipage}{3cm} 
\hspace{-2cm}\begin{picture}(50,50)(150,150)
\put(200,185){Cats}
\put(208,170){$n$}
\put(230,185){like}
\put(228,170){$n^r$}
\put(240,170){$s$}
\put(248,170){$n^l$}
\put(260,185){milk.}
\put(270,170){$n$}
\end{picture}

\vspace{-0.5cm}
\hspace{-0.15cm}{\epsfig{figure=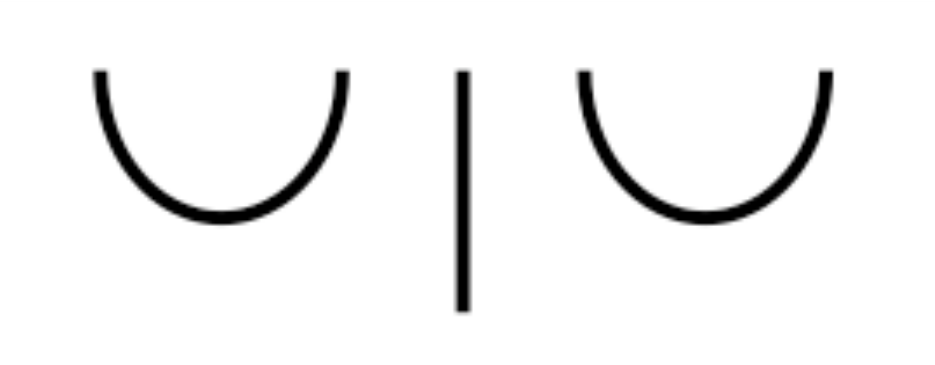,width=80pt}} 
\end{minipage}
\end{center}
\caption{The pregroup types and reduction diagram for a transitive sentence. }
\label{fig:pregroup}
\end{figure}

In a nutshell, pregroup types are either atomic or compound. Atomic types can be simple (\emph{e.g.}~$n$ for noun phrases, $s$ for statements) or left/right superscripted---referred to as adjoint types (\emph{e.g.}~$n^r$ and $n^l$). An example of a compound type is that of a verb  $n^r s n^l$.  The superscripted types  express that the verb is a relation with two arguments of type $n$, which have to occur to the \underline{\emph{r}}ight and to the \underline{\emph{l}}eft of it, and that it outputs an argument of the type $s$. A transitive sentence has  types as shown in Figure~\ref{fig:pregroup}.  

Each type $n$ cancels out with its right adjoint $n^r$ from the right and its left adjoint $n^l$ from the left; mathematically speaking these mean\footnote{The relation $\leq$ is the partial order of the pregroup. It  corresponds to implication $\implies$ in a logical reading thereof. If these inequalities are replaced by equalities, \emph{i.e.} if $n^ln=1=nn^r$, then the pregroup collapses into a group where $n^l = n^r$.} 
\[
n^ln\leq 1 \qquad \text{and} \qquad  nn^r \leq 1
\]
Here   1 is the unit of concatenation: $1n=n1=n$. The corresponding grammatical reduction of a transitive sentence  is $nn^rsn^l \leq 1s1 = s$. Each such reduction can be depicted as a wire diagram. The diagram of a transitive sentence is shown in Figure~\ref{fig:pregroup}.

\vspace{-2mm}

\vspace{-2mm}
\paragraph{Syntax-guided Semantic Composition}

According to~\newcite{Coeckeetal} and based on a general completeness theorem between compact categories, wire diagrams, and vector spaces, the  meaning of sentences can be canonically reduced to linear algebraic formulae. The following is the meaning vector of our transitive sentence:
\[
\overrightarrow{\text{cats} \ \text{like} \ \text{milk}}  =  
(f)\!\left(\overrightarrow{\text{cats}} \otimes \overrightarrow{\text{like}} \otimes \overrightarrow{\text{milk}}\right) \qquad {\bf (I)}
\]  
Here $f$ is the  linear map  that encodes the grammatical structure. The categorical morphism corresponding to it is denoted by the tensor product of 3 components: $\epsilon_V \otimes 1_S \otimes \epsilon_W$, where $V$ and $W$ are subject and object spaces, $S$ is the sentence space,  the $\epsilon$'s are the cups, and $1_S$ is the straight line in the diagram.   The cups stand for taking inner products, which when done with the basis vectors imitate substitution. The straight line stands for the identity map that does nothing. By the rules of the category,  equation {\bf (I)}  reduces to  the following  linear algebraic formula with lower dimensions, hence the dimensional explosion problem for Kronecker products is avoided:

\[
{\sum_{itj} c_{itj} \langle \overrightarrow{\text{cats}}\!\! \mid \!\! \overrightarrow{v_i}\rangle  \overrightarrow{s_t} 
\langle  \overrightarrow{w_j}\!\!\mid \!\!\overrightarrow{\text{milk}}\rangle\in S} \qquad {\bf (II)}
\]


\noindent
$\overrightarrow{v_i}, \overrightarrow{w_j}$ are basis vectors of $V$ and $W$. The inner product  $\langle \overrightarrow{\text{cats}}\!\! \mid \!\! \overrightarrow{v_i}\rangle $ substitutes the weights of $\overrightarrow{\text{cats}}$ into the first argument place of the verb (similarly for object and second argument place).   $\overrightarrow{s_t}$ is a basis vector of the sentence space $S$ in which meanings  of sentences  live, regardless of their grammatical structure.  

The  degree of synonymity of sentences is  obtained by taking the cosine measure of  their vectors.    $S$ is an abstract space: it needs to be instantiated to provide concrete meanings and synonymity  measures. For instance, a truth-theoretic model is obtained by taking the sentence space  $S$  to be the 2-dimensional space with basis vectors $\mid\!\!1\rangle$ (True) and $\mid\!\!0\rangle$ (False). 


\section{Building  Matrices for Relational Words}
\label{sec:building}
In this section we present a general scheme to build matrices for relational words.  Recall that given a vector space $A$ with basis $\{\overrightarrow{n_i}\}_i$,   the Kronecker product of two vectors $\overrightarrow{v} = \sum_i{c^{a}_i \overrightarrow{n_i}}$ and $\overrightarrow{w} = \sum_i{c^{b}_i \overrightarrow{n_i}}$  is defined as follows:

\[
\overrightarrow{v} \otimes \overrightarrow{w} = \sum_{ij}{c^a_i c^b_j\,  (\overrightarrow{n_i}\otimes\overrightarrow{n_j}})
\]

\noindent
where $(\overrightarrow{n_i}\otimes\overrightarrow{n_j})$ is just the pairing of the basis of $A$, \emph{i.e.} $(\overrightarrow{n_i}, \overrightarrow{n_j})$.  The Kronecker product vectors belong in the tensor product of $A$ with itself: $A \otimes A$, hence if $A$ has dimension $r$, these will be of dimensionality $r \times r$. The point-wise multiplication of these vectors is defined as follows

 \[
 \overrightarrow{v} \odot \overrightarrow{w} = \sum_i{c^a_i c^b_i\, \overrightarrow{n_i}} 
 \]

The intuition behind having a matrix for a relational word is that any relation $R$ on sets $X$ and $Y$, \emph{i.e.}~$R \subseteq X \times Y$ can be represented as a matrix, namely one that has as row-bases $x \in X$ and as column-bases $y \in Y$, with weight $c_{xy} = 1$ where $(x,y) \in R$ and 0 otherwise. In a distributional setting, the weights, which are natural or real numbers, will represent more: `the extent according to which $x$ and $y$ are related'. This can be determined in different ways. 

Suppose $X$ is the set of animals, and `chase' is a relation on it:  $chase \subseteq X \times X$.  Take $x =$ `dog' and $y=$ `cat': with  our type-logical glasses on, the obvious choice would be to take $c_{xy}$ to be the number of times `dog' has chased `cat', \emph{i.e.}~the number of times the sentence `the dog chases the cat' has appeared in the corpus. But in the distributional setting, this method will be too syntactic and dismissive  of the actual \emph{meaning} of `cat' and `dog'.  If instead the corpus contains the sentence `the hound hunted the wild cat', $c_{xy}$ will be 0, restricting us to only assign meaning to sentences that have directly appeared in the corpus. We propose to, instead, use a level of abstraction by taking words such as verbs to be distributions over the semantic information in the vectors of their context words, rather than over the context words themselves.

Start with an $r$-dimensional  vector space $N$ with basis $\left \{\overrightarrow{n}_i\right\}_i$, in which meaning vectors of atomic words, such as nouns, live.  The basis vectors of $N$ are in principle all  the words from the corpus, however in practice and following~\newcite{Lapata} we had to restrict these to a subset of  the most occurring words. These basis vectors are not restricted to nouns: they can as well be verbs, adjectives, and adverbs, so that we can define the meaning of a noun in all possible contexts---as is usual in context-based models---and not only in the context of other nouns.  Note that basis words with relational types are treated as pure lexical items rather than as semantic objects represented as matrices. In short, we count how many times a noun has occurred close to words of other syntactic types such as `elect' and  `scientific', rather than count how many times it has occurred close to their corresponding matrices: it is the lexical tokens that form the context, not their meaning.

Each relational word $P$ with grammatical type $\pi$ and $m$ adjoint types $\alpha_1, \alpha_2, \cdots, \alpha_m$  is encoded as an  $({r\times \ldots \times r})$ matrix with $m$ dimensions. Since our vector space $N$ has a fixed basis, each such matrix is represented in vector form as follows:
\[
\overrightarrow{\text{P}} = \sum_{\underbrace{ij \cdots \zeta}_{m}} c_{ij \cdots  \zeta} \ \underbrace{(\overrightarrow{n}_i\otimes \overrightarrow{n}_j \otimes \cdots \otimes \overrightarrow{n}_\zeta)}_{m}
\]
This vector lives in the tensor space $\underbrace{N \otimes N \otimes \cdots \otimes N}_m$. Each $c_{ij \cdots  \zeta}$ is computed according to the procedure described in Figure~\ref{fig:gen-matrix}. 

\begin{figure}[h!]
\fbox{\begin{minipage}{7.5cm}
{\bf 1)}\ Consider a sequence of words containing a relational word `P' and its arguments w$_1$, w$_2$, $\cdots$, w$_m$, occurring in the same order as described in P's grammatical type $\pi$. Refer to these sequences as `P'-relations. Suppose there are $k$ of them.\\
{\bf 2)}  Retrieve the vector $\overrightarrow{\text{w}}_l$ of each argument w$_l$.\\
{\bf 3)}\ Suppose  w$_1$ has weight $c^1_i$ on basis vector $\overrightarrow{n}_i$, w$_2$ has weight $c^2_j$ on basis vector $\overrightarrow{n}_j$,  $\cdots$, and  w$_m$ has weight $c^m_\zeta$ on basis vector $\overrightarrow{n}_\zeta$.  Multiply   these weights

\[
c^1_i \times c^2_j \times \cdots \times c^m_\zeta
\]

{\bf 4)} Repeat the above steps for all the $k$ `P'-relations, and sum\footnote{We also experimented with multiplication, but the sparsity of noun vectors resulted in most verb matrices being empty.} the corresponding weights

\[
c_{ij \cdots  \zeta} = \sum_k \left(c^1_i \times c^2_j \times \cdots \times c^m_\zeta \right)_k
\]
\end{minipage}}
\caption{Procedure for learning  weights for matrices of words `P' with relational types $\pi$ of $m$ arguments.}
\label{fig:gen-matrix}
\end{figure}

\medskip
\noindent
Linear algebraically, this procedure corresponds to computing the following 
\[
\overrightarrow{\text{P}} = \sum_k \left(\overrightarrow{\text{w}}_1 \otimes \overrightarrow{\text{w}}_2 \otimes \cdots \otimes \overrightarrow{\text{w}}_m
\right)_k
\]

Type-logical examples of relational  words are verbs, adjectives, and adverbs. A transitive verb is represented as a 2 dimensional matrix since its type is $n^rsn^l$ with two adjoint types $n^r$ and $n^l$.  
The corresponding vector of this matrix  is

\[
\overrightarrow{\text{verb}} = \sum_{ij} c_{ij}\, (\overrightarrow{n}_i\otimes  \overrightarrow{n}_j)
\]

The  weight $c_{ij}$  corresponding to basis vector $\overrightarrow{n}_i \otimes \overrightarrow{n}_j$,  is  the extent according to which words  that have co-occurred with  $\overrightarrow{n}_i$ have been the subject of  the `{verb}' and  words that have co-occurred with $\overrightarrow{n}_j$  have been the object of the `verb'. This example computation is demonstrated in Figure~\ref{fig:verb-matrix}.

\begin{figure}[h]
\fbox{\begin{minipage}{7.5cm}
{\bf 1)} Consider phrases containing `{verb}', its subject w$_1$ and object w$_2$. Suppose there are $k$ of them. \\
{\bf 2)}  Retrieve vectors $\overrightarrow{\text{w}}_1$ and $\overrightarrow{\text{w}}_2$. \\
{\bf 3)} Suppose $\overrightarrow{\text{w}}_1$ has weight $c^1_i$ on $\overrightarrow{n}_i$ and $\overrightarrow{\text{w}}_2$ has $c^2_j$ on $\overrightarrow{n}_j$.  Multiply these weights $
c^1_i \times c^2_j$. 

\noindent
{\bf 4)} Repeat the above steps for all $k$ `verb'-relations and  sum  the corresponding weights $\sum_{k} (c^1_i \times c^2_j)_k$
\end{minipage}}
\caption{Procedure for learning  weights for matrices of transitive verbs.}
\label{fig:verb-matrix}
\end{figure}

\medskip
\noindent
Linear algebraically, we are computing

\[\overrightarrow{\text{verb}} \quad =\quad
\sum_{k} \left(\overrightarrow{\text{w}}_1 \otimes \overrightarrow{\text{w}}_2\right)_k
\]

\noindent
As an example, consider the verb `show' and suppose there are two `{show}'-relations  in the corpus:

\vspace{0mm}
\begin{tabular}{ccll}
$s_1$ &\quad = \quad & table show result\\
$s_2$ & \quad = \quad & map show location
\end{tabular}

\smallskip
\noindent
The vector of `show' is

\[\overrightarrow{\text{show}} \ = \  
\overrightarrow{\text{table}} \otimes \overrightarrow{\text{result}} \ + \
\overrightarrow{\text{map}} \otimes \overrightarrow{\text{location}}
\]

\noindent
Consider an $N$ space with four basis vectors `far', `room', `scientific', and `elect'. The TF/IDF-weighted values for vectors of the above four nouns (built from the BNC) are as shown in Table \ref{tab:sample_weights}.

\begin{table}[h!]
	\begin{center}
		\begin{tabular}{c|c|c|c|c|c}
		\hline
		$\mathbf{i}$ & $\overrightarrow{\mathbf{n_i}}$ &  table & map & result & location \\
		\hline
		\hline

		1&far &6.6 & 5.6 &7& 5.9\\

		2&room &  27 & 7.4 &0.99& 7.3\\

		3&scientific& 0& 5.4& 13& 6.1\\

		4&elect & 0  & 0&4.2&0
		\end{tabular}
	\end{center}
	\caption{Sample  weights for selected noun vectors.}
	\label{tab:sample_weights}
\end{table}

\noindent
Part of the matrix of `show' is presented in Table \ref{tab:show_matrix}.

\begin{table}[h]
	\begin{center}
		\begin{tabular}{c|c|c|c|c}
		\hline

		&far & room & scientific & elect\\
		\hline

		far &\fbox{79.24} &47.41&119.96&27.72\\
		room& 232.66 & 80.75&396.14& 113.2\\
		scientific &32.94&31.86 &32.94&0\\
		elect&0&0&0&0\\

		\end{tabular}
		\end{center}
	\caption{Sample semantic matrix for  `show'.}
	\label{tab:show_matrix}
\end{table}
 
 \medskip
 \noindent
As a sample computation, the weight  $c_{11}$ for \\
 vector $(1,1)$, \emph{i.e.}~$(\overrightarrow{\text{far}}, \overrightarrow{\text{far}})$ is computed by multiplying weights of `table' and `result' on $\overrightarrow{\text{far}}$,  \emph{i.e.}~$6.6 \times 7$, multiplying weights of `map' and `location' on $\overrightarrow{\text{far}}$, \emph{i.e.}~$5.6 \times 5.9$ then adding these $46.2 + 33.04$ and obtaining the total weight $79.24$.

\medskip
\noindent
The same method is applied to build matrices for ditransitive verbs, which will have 3 dimensions, and adjectives and adverbs, which will be of 1 dimension each.

\section{Computing  Sentence Vectors}
Meaning of sentences are vectors computed by taking the variables of the  categorical prescription of meaning (the linear map $f$ obtained from the grammatical reduction of the sentence)  to be determined by the matrices of the relational words. For instance the meaning of the transitive sentence `sub {verb} obj'  is:
\[
\overrightarrow{\text{sub} \ \text{verb} \ \text{obj}} = \sum_{itj}  \langle \overrightarrow{\text{sub}} \mid \overrightarrow{v}_i \rangle  \langle \overrightarrow{w}_j \mid \overrightarrow{\text{obj}}\rangle\, c_{itj} \overrightarrow{s}_t
\]
We take  $V := W := N$ and $S = N \otimes N$,  then $\sum_{itj}c_{itj}\overrightarrow{s}_t$ is  determined by the matrix of the verb, \emph{i.e.} substitute it by  $\sum_{ij} c_{ij} (\overrightarrow{n}_i \otimes \overrightarrow{n}_j)$\footnote{Note that by doing so we are also reducing the  verb space from $N \otimes (N \otimes N) \otimes N$ to $N \otimes N$, since for our construction we only need tuples of the form $\overrightarrow{n}_i \otimes \overrightarrow{n}_i \otimes \overrightarrow{n}_j \otimes \overrightarrow{n}_j$  which are isomorphic to pairs $(\overrightarrow{n}_i \otimes \overrightarrow{n}_j)$.}. Hence   $\overrightarrow{\text{sub} \ \text{verb} \ \text{obj}}$ becomes:
\begin{eqnarray*}
 \sum_{ij} \langle \overrightarrow{\text{sub}} \mid \overrightarrow{n}_i \rangle \langle \overrightarrow{n}_j \mid \overrightarrow{\text{obj}}  \rangle  c_{ij} (\overrightarrow{n}_i \otimes \overrightarrow{n}_j) &=&\\
\sum_{ij} c^{sub}_i c^{obj}_j   c_{ij} (\overrightarrow{n}_i \otimes \overrightarrow{n}_j)
\end{eqnarray*}
This  can be decomposed to point-wise multiplication of two vectors as follows:
\[
\Big(\sum_{ij} c^{sub}_i c^{obj}_j (\overrightarrow{n}_i\otimes  \overrightarrow{n}_j)\Big) \ \odot \ \Big(\sum_{ij}c_{ij} (\overrightarrow{n}_i \otimes \overrightarrow{n}_j)\Big)
\]
The left argument is the Kronecker product of subject and object vectors and the right argument is the vector of the verb, so we  obtain 
\[
\left(\overrightarrow{\text{sub}} \otimes \overrightarrow{\text{obj}} \right) \odot \overrightarrow{\text{verb}}
\]
Since $\odot$ is commutative, this provides us with a distributional version of the type-logical meaning of the sentence: point-wise multiplication of the meaning of the verb to the Kronecker product of its subject and object: 
\[
\overrightarrow{\text{sub} \ \text{verb} \ \text{obj}} \quad = \quad \overrightarrow{\text{verb}} \odot \left(\overrightarrow{\text{sub}} \otimes  \overrightarrow{\text{obj}} \right)
\]
This mathematical operation can be informally described as a structured `mixing' of the information of the subject and object, followed by it being `filtered' through the information of the verb applied to them, in order to produce the information of the sentence.

In the transitive case,  $S = N \otimes N$, hence $\overrightarrow{s}_t = \overrightarrow{n}_i \otimes \overrightarrow{n}_j$. More generally,   the vector space corresponding to  the abstract sentence space $S$ is the concrete tensor space  $({N\otimes \ldots \otimes N})$ for $m$  the dimension of the matrix of the `verb'. As we have seen above, in practice we do not need to build this tensor space, as the computations thereof reduce to point-wise multiplications and summations. 

Similar computations yield meanings of sentences with adjectives and adverbs. For instance the meaning of a transitive sentence with a modified subject and a modified verb  we have 
\begin{align*}
 & \overrightarrow{\mbox{adj  sub verb obj adv}} = & \\
 & \left(\overrightarrow{\text{adv}} \odot \overrightarrow{\text{verb}}\right) \odot \left(\left(\overrightarrow{\text{adj}}\odot \overrightarrow{\text{sub}}\right) \otimes \overrightarrow{\text{obj}}\right) &
 \end{align*}
After building vectors for sentences, we can compare their meaning and measure their degree of synonymy by taking their cosine measure. 

\section{Evaluation}

Evaluating such a framework is no easy task. What to evaluate depends heavily on what sort of application a practical instantiation of the model is geared towards. In~\cite{Grefenetal}, it is suggested that the simplified model we presented and expanded here could be evaluated in the same way as lexical semantic models, measuring compositionally built sentence vectors against a benchmark dataset such as that provided by \newcite{Lapata}. In this section, we briefly describe the evaluation of our model against this dataset. Following this, we present a new evaluation task extending the experimental methodology of \newcite{Lapata} to transitive verb-centric sentences, and compare our model to those discussed by \newcite{Lapata} within this new experiment.

\paragraph{First Dataset Description}

The first experiment, described in detail by \newcite{Lapata}, evaluates how well compositional models disambiguate ambiguous words given the context of a potentially disambiguating noun. Each entry of the dataset provides a noun, a target verb and landmark verb (both intransitive). The noun must be composed with both verbs to produce short phrase vectors the similarity of which is measured by the candidate. Also provided with each entry is a classification (``High'' or ``Low'') indicating whether or not the verbs are indeed semantically close within the context of the noun, as well as an evaluator-set similarity score between 1 and 7 (along with an evaluator identifier), where 1 is low similarity and 7 is high.

\paragraph{Evaluation Methodology}

Candidate models provide a similarity score for each entry. The scores of high similarity entries and low similarity entries are averaged to produce a mean High score and mean Low score for the model. The correlation of the model's similarity judgements with the human judgements is also calculated using Spearman's $\rho$, a metric which is deemed to be more scrupulous, and ultimately that by which models should be ranked, by \newcite{Lapata}. The mean for each model is on a $[0,1]$ scale, except for UpperBound which is on the same $[1,7]$ scale the annotators used. The $\rho$ scores are on a $[-1,1]$ scale. It is assumed that inter-annotator agreement provides the theoretical maximum $\rho$ for any model for this experiment. The cosine measure of the verb vectors, ignoring the noun, is taken to be the baseline (no composition).

\paragraph{Other Models}

The other models we compare ours to are those evaluated by \newcite{Lapata}. We provide a selection of the results from that paper for the worst (Add) and best\footnote{The multiplicative model presented here is what is qualified as best in~\cite{Lapata}. However, they also present a slightly better performing ($\rho=0.19$) model which is a combination of their multiplicative model and a weighted additive model. The difference in $\rho$ is qualified as ``not statistically significant'' in the original paper, and furthermore the mixed model requires parametric optimisation hence was not evaluated against the entire test set. For these reasons, we chose not to include it in the comparison.} (Multiply) performing models, as well as the previous second-best performing model (Kintsch). The additive and multiplicative models are simply applications of vector addition and component-wise multiplication. We invite the reader to consult~\cite{Lapata} for the description of Kintsch's additive model and parametric choices.

\paragraph{Model Parameters}
To provide the most accurate comparison with the existing multiplicative model, and exploiting the aforementioned feature that the categorical model can be built ``on top of'' existing lexical distributional models, we used the parameters described by \newcite{Lapata} to reproduce the vectors evaluated in the original experiment as our noun vectors. All vectors were built from a lemmatised version of the BNC. The noun basis was the 2000 most common context words, basis weights were the probability of context words given the target word divided by the overall probability of the context word. Intransitive verb function-vectors were trained using the procedure presented in $\S$\ref{sec:building}. Since the dataset only contains intransitive verbs and nouns, we used $S=N$. The cosine measure of vectors was used as a similarity metric.

\paragraph{First Experiment Results}

In Table \ref{tab:results} we present the comparison of the selected models. Our categorical model performs significantly better than the existing second-place (Kintsch) and obtains a $\rho$ quasi-identical to the multiplicative model, indicating significant correlation with the annotator scores.

\begin{table}[h]
\begin{center}
\begin{tabular}{|llll|}
\hline
Model & High & Low & $\rho$\\
\hline
\hline
Baseline & 0.27 & 0.26 & 0.08\\
\hline
\hline
Add & 0.59 & 0.59 & 0.04\\
Kintsch & 0.47 & 0.45 & 0.09\\
Multiply & 0.42 & 0.28 & 0.17 \\
\textbf{Categorical} & \textbf{0.84} & \textbf{0.79} & \textbf{0.17}\\
\hline
\hline
UpperBound & 4.94 & 3.25 & 0.40 \\
\hline
\end{tabular}
\end{center}
\caption{Selected model means for High and Low similarity items and correlation coefficients with human judgements, first experiment~\cite{Lapata}. $p < 0.05$ for each $\rho$.}
\label{tab:results}
\end{table}

\begin{figure}[h]
\hspace{-0.15cm}{\epsfig{figure=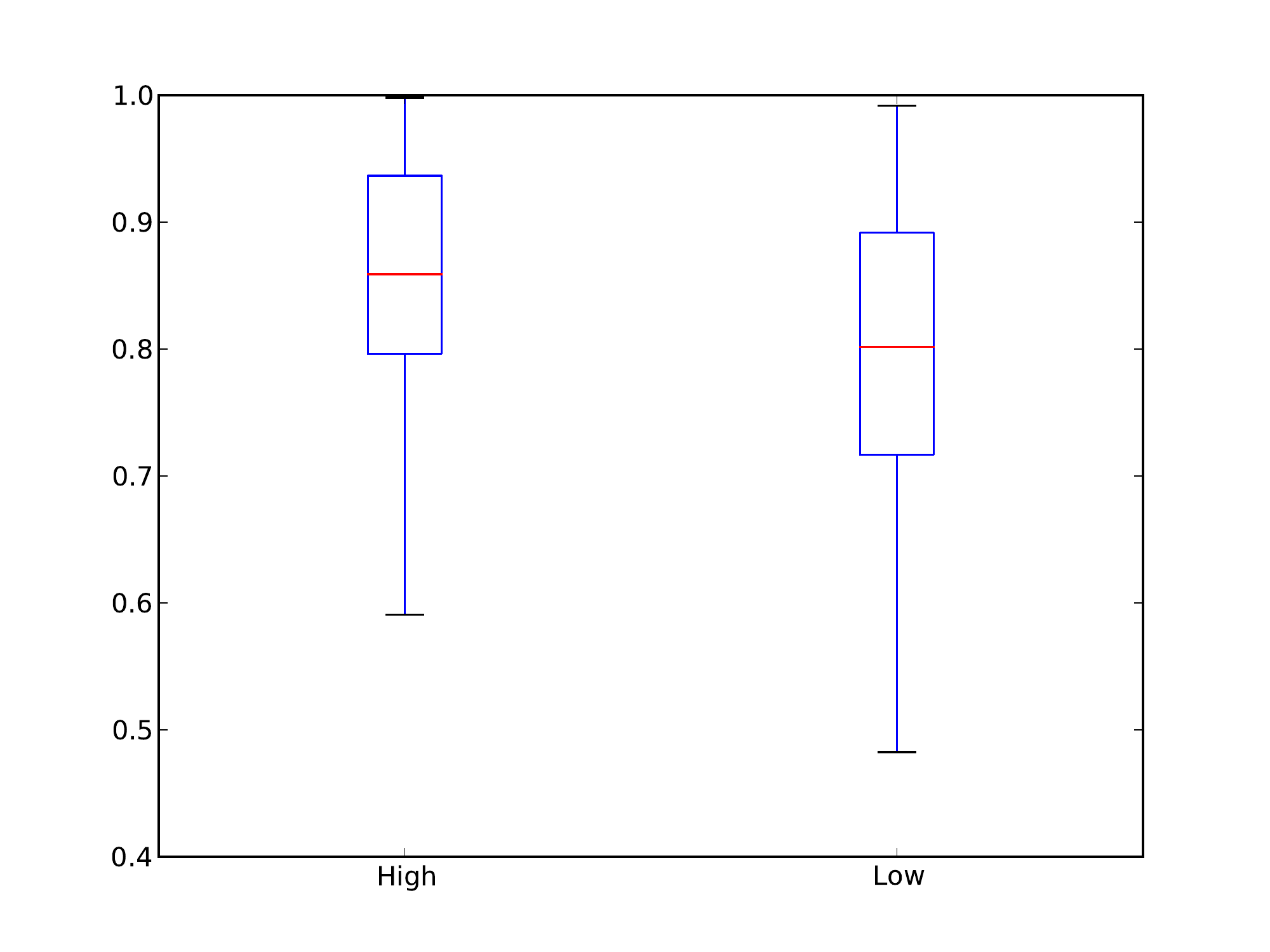,width=230pt}}
\caption{Distribution of predicted similarities for the categorical distributional model on High and Low similarity items.}
\label{fig:score_dist}
\end{figure}

There is not a large difference between the mean High score and mean Low score, but the distribution in Figure \ref{fig:score_dist} shows that our model makes a non-negligible distinction between high similarity phrases and low similarity phrases, despite the absolute scores not being different by more than a few percentiles.

\paragraph{Second Dataset Description}

The second dataset\footnote{\texttt{http://www.cs.ox.ac.uk/activities/CompD\\istMeaning/GS2011data.txt}}, developed by the authors, follows the format of the \cite{Lapata} dataset used for the first experiment, with the exception that the target and landmark verbs are transitive, and an object noun is provided in addition to the subject noun, hence forming a small transitive sentence. The dataset comprises 200 entries consisting of sentence pairs (hence a total of 400 sentences) constructed by following the procedure outlined in $\S$4 of \cite{Lapata}, using transitive verbs from CELEX\footnote{\texttt{http://celex.mpi.nl/}}. For examples of these sentences, see Table~\ref{tab:trans-sent}. The dataset was split into four sections of 100 entries each, with guaranteed 50\% exclusive overlap with exactly two other datasets. Each section was given to a group of evaluators, with a total of 25, who were asked to form simple transitive sentence pairs from the verbs, subject and object provided in each entry; for instance  `the table showed the result' from  `table show result'. The evaluators were then asked to rate the semantic similarity of each verb pair within the context of those sentences, and offer a score between 1 and 7 for each entry. Each entry was given an arbitrary classification of HIGH or LOW by the authors, for the purpose of calculating mean high/low scores for each model. For example, the first two pairs in table~\ref{tab:trans-sent} were classified as HIGH, whereas the second two pairs as LOW.

\begin{table}[h]
\begin{center}
 \begin{tabular}{|c|c|}
 \hline
 Sentence 1 & Sentence 2\\
 \hline
 \hline
 
 table show result & table express  result  \\
\hline
 map show location & map picture location\\
 \hline
  table show result & table picture result\\
\hline
 map show location & map express location\\
 \hline
 \end{tabular}
\end{center}
\caption{Example entries from the transitive dataset without annotator score, second experiment.}
\label{tab:trans-sent}
\end{table}

\paragraph{Evaluation Methodology}

The evaluation methodology for the second experiment was identical to that of the first, as are the scales for means and scores. Here also, Spearman's $\rho$ is deemed a more rigorous way of determining how well a model tracks difference in meaning. This is both because of the imprecise nature of the classification of verb pairs as HIGH or LOW; and since the objective similarity scores produced by a model that distinguishes sentences of different meaning from those of similar meaning can be renormalised in practice. Therefore the delta between HIGH means and LOW mean cannot serve as a definite indication of the practical applicability (or lack thereof) of semantic models; the means are provided just to aid comparison with the results of the first experiment.

\paragraph{Model Parameters}

As in the first experiment, the lexical vectors from \cite{Lapata} were used for the other models evaluated (additive, multiplicative and baseline)\footnote{Kintsch was not evaluated as it required optimising model parameters against a held-out segment of the test set, and we could not replicate the methodology of \newcite{Lapata} with full confidence.} and for the noun vectors of our categorical model. Transitive verb vectors were trained as described in $\S$\ref{sec:building} with $S = N \otimes N$.

\paragraph{Second Experiment Results}

The results for the models evaluated against the second dataset are presented in Table $\ref{tab:results_2}$.

\begin{table}[h]
\begin{center}
\begin{tabular}{|llll|}
\hline
Model & High & Low & $\rho$\\
\hline
\hline
Baseline & 0.47 & 0.44 & 0.16\\
\hline
\hline
Add & 0.90 & 0.90 & 0.05\\
Multiply & 0.67 & 0.59 & 0.17 \\
\textbf{Categorical} & \textbf{0.73} & \textbf{0.72} & \textbf{0.21}\\
\hline
\hline
UpperBound & 4.80 & 2.49 & 0.62 \\
\hline
\end{tabular}
\end{center}
\caption{Selected model means for High and Low similarity items and correlation coefficients with human judgements, second experiment. $p < 0.05$ for each $\rho$.}
\label{tab:results_2}
\end{table}

\noindent
We observe a significant (according to $p < 0.0.5$) improvement in the alignment of our categorical model with the human judgements, from 0.17 to 0.21. The additive model continues to make little distinction between senses of the verb during composition, and the multiplicative model's alignment does not change, but becomes statistically indistinguishable from the non-compositional baseline model. 

Once again we note that the high-low means are not very indicative of model performance, as the difference between high mean and the low mean of the categorical model is much smaller than that of the both the baseline model and multiplicative model, despite better alignment with annotator judgements.


\section{Discussion}

In this paper, we described an implementation of the categorical model of meaning~\cite{Coeckeetal}, which combines the formal logical and the empirical distributional frameworks into a unified semantic model.  The implementation is based on  building matrices for words with relational types (adjectives, verbs), and vectors for words with atomic types (nouns), based on data from the BNC. We then show how to apply verbs to their subject/object, in order to compute the meaning of intransitive and transitive sentences. 

Other work uses matrices to model meaning~\cite{Baroni,Guevara}, but only for  adjective-noun phrases.  Our approach easily applies to such compositions, as well as to sentences containing combinations of adjectives, nouns, verbs, and adverbs. The other key difference is that they learn their matrices in a top-down fashion, \emph{i.e.} by regression from the composite adjective-noun context vectors, whereas our model is bottom-up: it learns sentence/phrase meaning compositionally from the vectors of the compartments of the composites. Finally, very similar functions, for example a verb with argument alternations such as `break' in `Y breaks' and `X breaks Y', are not treated as unrelated. The matrix of the intransitive `break' uses the corpus-observed information about the subject of break, including that of `Y', similarly the matrix of the transitive `break' uses information about its subject and object, including that of `X' and `Y'. We leave a thorough study of these phenomena, which fall under providing a modular representation of  passive-active similarities, to future work. 

We evaluated our model in two ways: first against the word disambiguation task of~\newcite{Lapata} for intransitive verbs, and then against a similar new experiment  for  transitive verbs, which we developed.

Our findings in the first experiment show that the categorical method performs on par with the leading existing approaches. This should not surprise us given that the context is so small and our method becomes similar to the multiplicative model of \newcite{Lapata}. However, our approach is sensitive to grammatical structure, leading us to develop a second experiment taking this into account and differentiating it from models with commutative composition operations.
 
  The second experiment's results deliver the expected qualitative difference between models, with our categorical model outperforming the others and showing an increase in alignment with human judgements in correlation with the increase in sentence complexity. 
We use this second evaluation principally to show that there is a strong case for the development of more complex experiments measuring not only the disambiguating qualities of compositional models, but also their syntactic sensitivity, which is not directly measured in the existing experiments. 

These results show that the high level categorical distributional model, uniting empirical data with logical form, can be implemented just like any other concrete model. Furthermore it shows better results in experiments involving higher syntactic complexity. This is just the tip of the iceberg: the mathematics underlying the implementation ensures that it uniformly scales to larger, more complicated sentences and enables it to compare synonymity of sentences that are of different grammatical structure.

\section{Future Work}

Treatment of function words such as `that', `who',  as well as logical words such as quantifiers and conjunctives are left to future work. This will build alongside  the general guidelines of \newcite{Coeckeetal} and concrete insights from the work of~\newcite{Widdows}. It is not yet entirely clear how existing set-theoretic approaches, for example that of discourse representation and generalised quantifiers,  apply to our setting. Preliminary work on integration of the two has been  presented by~\newcite{Preller} and more recently also by~\newcite{PrellerSadr}. 

As mentioned by one of the reviewers, our pregroup approach to grammar flattens the sentence representation, in that the verb is applied to its subject and object at the same time; whereas in other approaches such as CCG, it is first applied to the object to produce a verb phrase, then applied to the subject to produce the sentence. The advantages and disadvantages of this method and comparisons with other systems, in particular CCG, constitutes ongoing work. 

\section{Acknowledgement}
We wish to thank P.~Blunsom, S.~Clark, B.~Coecke, S.~Pulman, and the anonymous EMNLP reviewers for discussions and comments. Support from EPSRC grant EP/F042728/1 is gratefully acknowledged by M.~Sadrzadeh.


\begin{thebibliography}{}
	

\bibitem[\protect\citename{Alshawi}1992]{Alshawi}
 H.~Alshawi (ed).
 \newblock 1992.
 \newblock \emph{The Core Language Engine}.
 \newblock MIT Press.

\bibitem[\protect\citename{Baroni and Zamparelli}2010]{Baroni}
 M.~Baroni and R.~Zamparelli.
 \newblock 2010.
 \newblock \emph{Nouns are vectors, adjectives are matrices}.
 \newblock Proceedings of {Conference on Empirical Methods in Natural Language Processing (EMNLP)}. 



\bibitem[\protect\citename{Clark and Pulman}2007]{ClarkPulman}
S.~Clark and S.~Pulman.
\newblock 2007.
\newblock \emph{Combining Symbolic and Distributional Models of Meaning}.
\newblock Proceedings of AAAI Spring Symposium on {Q}uantum {I}nteraction.
\newblock  {AAAI Press}.

\bibitem[\protect\citename{Coecke and Paquette}2011]{Coecke}
B.~Coecke, and E.~Paquette.
\newblock 2011.
\newblock \emph{Categories for the Practicing Physicist}. 
\newblock  \emph{New Structures for Physics}, 167Ð-271. 
\newblock B.~Coecke (ed.).
\newblock Lecture Notes in Physics {\bf 813}.
\newblock Springer.

\bibitem[\protect\citename{Coecke et al.}2010]{Coeckeetal}  
B.~Coecke, M.~Sadrzadeh and S.~Clark.
\newblock 2010.
\newblock  \emph{Mathematical Foundations for Distributed  Compositional Model of Meaning}.
\newblock Lambek Festschrift.
\newblock   {Linguistic Analysis} {\bf 36}, 345--384.
\newblock  J. van Benthem, M. Moortgat and W. Buszkowski (eds.). 


\bibitem[\protect\citename{Curran}2004]{Curran}
 J.~Curran.
 \newblock 2004.
 \newblock \emph{From Distributional to Semantic Similarity}.
 \newblock PhD Thesis, University of Edinburgh.
	
\bibitem[\protect\citename{Erk and Pado}2008]{ErkPado}
 K.~Erk and S.~Pad{\'o}.
 \newblock 2004.
 \newblock  \emph{A Structured Vector Space Model for Word Meaning in Context}.
\newblock Proceedings of {Conference on Empirical Methods in Natural Language Processing (EMNLP)}, 897--906. 

\bibitem[\protect\citename{Frege}1892]{Frege}
 G.~Frege
 \newblock 1892.
 \newblock  \emph{\"Uber Sinn und Bedeutung}.
 \newblock {Zeitschrift f\"ur Philosophie und philosophische Kritik 100}.

\bibitem[\protect\citename{Firth}1957]{Firth}
 J.~R.~Firth.
 \newblock 1957.
 \newblock  \emph{A synopsis of linguistic theory 1930-1955}.
 \newblock {Studies in Linguistic Analysis}. 

\bibitem[\protect\citename{Grefenstette et al.}2011]{Grefenetal} 
E.~Grefenstette, M.~Sadrzadeh, S.~Clark,  B.~Coecke, S.~Pulman.
\newblock 2011.
\newblock \emph{Concrete Compositional Sentence Spaces for a Compositional Distributional Model of Meaning}.
\newblock International Conference on Computational Semantics (IWCS'11).
\newblock Oxford. 

\bibitem[\protect\citename{Grefenstette}1994]{Grefenstette}
G.~Grefenstette.
\newblock 1994.
\newblock  \emph{Explorations in  Automatic Thesaurus Discovery}.
\newblock  Kluwer.

\bibitem[\protect\citename{Guevara}2010]{Guevara}
E.~Guevara. 
\newblock 2010.
\newblock \emph{A Regression Model of Adjective-Noun Compositionality in Distributional Semantics}.
\newblock Proceedings of the  ACL GEMS Workshop.


\bibitem[\protect\citename{Harris}1966]{Harris}
 Z.~S.~Harris.
 \newblock 1966.
 \newblock  \emph{A Cycling Cancellation-Automaton for Sentence Well-Formedness}.
\newblock {International Computation Centre Bulletin} {\bf 5}, 69--94.

\bibitem[\protect\citename{Hudson}1984]{Hudson}
 R.~Hudson.
 \newblock 1984.
 \newblock \emph{Word Grammar}.
 \newblock Blackwell. 
  
\bibitem[\protect\citename{Lambek}2008]{Lambek}
 J.~Lambek.
 \newblock 2008.
 \newblock  \emph{From Word to Sentence}.
 \newblock  Polimetrica, Milan. 

\bibitem[\protect\citename{Landauer}1997]{Landauer}
 T.~Landauer, and S.~Dumais.
 \newblock 2008.
 \newblock  \emph{A solution to Plato's problem: The latent semantic analysis theory of acquisition, induction, and representation of knowledge}.
 \newblock  Psychological review. 


\bibitem[\protect\citename{Manning et al.}2008]{Manning}
 C.~D.~Manning, P.~Raghavan, and H.~Sch{\"u}tze.
 \newblock 2008.
 \newblock  \emph{Introduction to information retrieval}.
 \newblock  Cambridge University Press. 

\bibitem[\protect\citename{Mitchell and Lapata}2008]{Lapata}
 J.~Mitchell and M.~Lapata.
 \newblock 2008.
 \newblock \emph{Vector-based models of semantic composition}.
 \newblock Proceedings of the {46th Annual Meeting of the Association for Computational Linguistics}, 236--244. 
	
\bibitem[\protect\citename{Montague}1974]{Montague}
 R.~Montague.
 \newblock 1974.
 \newblock \emph{English as a formal language}.
 \newblock {Formal Philosophy}, 189--223.
 
\bibitem[\protect\citename{Nivre}2003]{Nivre}
 J.~Nivre.
 \newblock 2003.
 \newblock \emph{An efficient algorithm for projective dependency parsing}.
 \newblock {Proceedings of the 8th International Workshop on Parsing Technologies (IWPT)}. 

\bibitem[\protect\citename{Preller}2007]{Preller}
A.~Preller.
\newblock \emph{Towards Discourse Representation via Pregroup Grammars}.
\newblock {Journal of Logic Language Information} {\bf 16} 173--194.



\bibitem[\protect\citename{Preller and Sadrzadeh} 2009]{PrellerSadr}
A.~Preller and M.~Sadrzadeh.
\newblock \emph{Semantic Vector Models and Functional Models for Pregroup Grammars}.
\newblock  Journal of Logic Language Information.
\newblock DOI: 10.1007/s10849-011-9132-2.
\newblock to appear. 

\bibitem[\protect\citename{Saffran et al.}1999]{Saffran}
 J.~Saffron, E.~Newport, R.~Asling.
 \newblock 1999.
 \newblock  \emph{Word Segmentation: The role of distributional cues}.
 \newblock {Journal of Memory and Language} {\bf 35}, 606--621.

\bibitem[\protect\citename{Schutze}1998]{Schutze}
  H.~Schuetze.
  \newblock 1998.
  \newblock \emph{Automatic Word Sense Discrimination}.
  \newblock  {Computational Linguistics} {\bf 24}, 97--123. 

\bibitem[\protect\citename{Smolensky}1990]{Smolensky}
  P.~Smolensky.
  \newblock 1990.
  \newblock \emph{Tensor product variable binding and the representation of symbolic structures in connectionist systems}.
  \newblock  {Computational Linguistics} {\bf 46, 1--2}, 159--216.

\bibitem[\protect\citename{Steedman}2000]{Steedman}
  M.~Steedman.
  \newblock 2000.
  \newblock \emph{The Syntactic Process}.
  \newblock  MIT Press.


\bibitem[\protect\citename{Widdows}2005]{Widdows}
  D.~Widdows.
  \newblock 2005.
  \newblock \emph{Geometry and Meaning}.
  \newblock  {University of Chicago Press}.

\bibitem[\protect\citename{Wittgenstein}1953]{Wittgenstein}
 L.~Wittgenstein.
 \newblock 1953.
 \newblock \emph{Philosophical Investigations}.
 \newblock Blackwell. 
	

\end{thebibliography}
\end{document}